\documentclass{article}
\usepackage{ijcai20}

\usepackage{times}

\usepackage{soul}
\usepackage{url}
\usepackage[hidelinks]{hyperref}
\usepackage[utf8]{inputenc}
\usepackage[small]{caption}
\usepackage{graphicx}
\usepackage{amsmath}
\usepackage{booktabs}
\usepackage{algorithm}
\usepackage{algorithmic}
\title{Heat and Blur: An Interpretability Based Defense Against Adversarial Examples}

\author{
Haya Brama\And
Tal Grinshpoun\footnote{Contact Author}\\
\affiliations
Department of Industrial Engineering and Management, Ariel University, Ariel, Israel\\
Ariel Cyber Innovation Center, Ariel University, Ariel, Israel\\
\emails
hayahartuv@gmail.com,
talgr@ariel.ac.il
}

\begin{document}

\maketitle

\begin{abstract}
The growing incorporation of artificial neural networks (NNs) into many fields, and especially into life-critical systems, is restrained by their vulnerability to adversarial examples (AEs).  Some existing defense methods can increase NNs' robustness, but they often require special architecture or training procedures and are irrelevant to already trained models. In this paper, we propose a simple defense that combines feature visualization with input modification, and can, therefore, be applicable to various pre-trained networks. By reviewing several interpretability methods, we gain new insights regarding the influence of AEs on NNs' computation. Based on that, we hypothesize that information about the ``true'' object is preserved within the NN’s activity, even when the input is adversarial, and present a feature visualization version that can extract that information in the form of relevance heatmaps. We then use these heatmaps as a basis for our defense, in which the adversarial effects are corrupted by massive blurring. We also provide a new evaluation metric that can capture the effects of both attacks and defenses more thoroughly and descriptively, and demonstrate the effectiveness of the defense and the utility of the suggested evaluation measurement with VGG19 results on the ImageNet dataset.
\end{abstract}

\section{Introduction}
\label{sec:intro}

Artificial neural networks (NNs) have been recently related to several breakthroughs in artificial intelligence tasks. However, their computational process usually involves millions of parameters and was initially considered inaccessible and uninterpretable to humans. This unintuitive nature of computation may lead to unpredicted (``Clever Hans'') mistakes, and may also give rise to different kinds of adversarial attacks. Such attacks produce adversarial examples (AEs) where small malicious changes, usually undetectable by humans, lead NNs to output wrong decisions with high confidence~\cite{szegedy2013intriguing}. In this work we suggest using visualization techniques, originally created as a tool for NN interpretability, in ways that: 1) provide a new perspective on the nature of AEs' impact; 2) can be utilized for a new and simple defense technique against adversarial attacks. Additionally, in order to evaluate our defense we introduce an evaluation metric that enables more informative analysis of the quality of both attacks and defenses.

\subsection{AE Explanations and Defenses}
\label{sec:ae}

Common explanations of the AE phenomenon focus on the failure of the models to map correctly the entire input space to the output space~\cite{tabacof2016exploring}. This can also be due to the linear nature of the computation, where small initial perturbations accumulate through the layers to cause significant deviations~\cite{goodfellow2014explaining}. On the other hand, other researchers put the blame on the non-linear processes or the high dimensionality of the input~\cite{bhagoji2017dimensionality}, while others argue that AEs occur where the decision boundary of the models lies close to the data manifold~\cite{tanay2016boundary}. A different approach relates the existence of AEs to non-robust features that exist in standard datasets~\cite{ilyas2019adversarial}.

Side by side with the attempts to explain AEs, many defense strategies have been proposed in order to increase the robustness of NNs against such attacks. Akhtar and Mian~\shortcite{akhtar2018threat} sorted those defenses along three main directions: 1) using modified training or modified input; 2) modifying the network by adding special layers or changing the loss/activation functions; and 3) using external models as network add-ons during classification. Though effective, the above methods as well as other methods do not provide a complete defense against AEs. Many of them have been evaluated with only a very specific attack type or under particular parameters, and fail when encountering different or stronger AEs~\cite{carlini2017towards}, or when considering the worst-case of limited or even perfect-knowledge attacks~\cite{carlini2017adversarial}. Moreover, many of the proposed defenses require extra computations and special training or architecture. Therefore even the relatively successful defenses may not be relevant to already trained models, or to tasks that require quick decisions or complex and sophisticated modeling. 

In the current research, we seek for a simple and fast method, in terms of both implementation and computational resources, to detect AEs. In order for this method to be applicable to different types of NNs, the defense should not require a special architecture or changes to the (already trained) network's parameters. A natural direction is to modify the input during testing, in a way that cancels-out the subtle adversarial perturbations. Previous such methods, like feature squeezing~\cite{xu2018feature}, have been shown to be highly successful in black-box scenarios where the attacker is unaware of the defense. However, they can be broken easily when the input-modification is taken into account while attacking~\cite{carlini2017adversarial}. Of course, the defender can respond by increasing the protective distortion level of the defense, but this will eventually lead to a decrease in the performance of the network in general, since benign inputs will become unrecognizable as well.

\subsection{Feature Visualization}
\label{sec:fv}

At this point we take advantage of the progress made in a parallel line of research, called feature (or network) visualization~\cite{olah2017feature}, which aims to understand NN computation and make NNs more interpretable. Specifically, several methods have been proposed for highlighting the input features which contribute to the network's decisions. Three main approaches to this end are: 1) input modification methods, which observe how changing the input to a NN affects the output; 2) class activation mapping methods, which combine activation patterns of a specific (usually high-level) layer with additional information such as the output and the gradients; and 3) back-propagation based methods, which are at the focus of the current work.

Back propagation based methods basically aim at reversing the inference process of the network. For example, Layer-wise Relevance Propagation (LRP)~\cite{bach2015pixel} propagates the final prediction backwards to the input in order to single-out the most influential components (e.g. pixels). Namely, the classification decision is decomposed into pixel-wise relevance scores, according to a conservation principle which forces the relevance quantity to be preserved between layers. Different reverse propagation rules have been suggested, as reflected in different variants of LRP such as LRP-$\epsilon$, LRP-$\alpha_1 \beta_0$ and LRP-$\gamma$~\cite{montavon2019layer}. However, the most basic attribution rule can be defined as:
\begin{equation}
R_i^{(l)}=\sum_j \frac{z_{ij}}{z_j} R_j^{(l+1)}
\label{eq2}
\end{equation}
where $z_{ij}= x_i w_{ij}$ and $R^{(l)}$ is the relevance of the $l$th layer. When restricted to the positive input spaces, this rule can be interpreted and theoretically explained as deep Tylor decomposition (DTD) of the ReLU activations~\cite{montavon2017explaining}.

Though robust against gradient shattering and resulting in fine-detailed heatmaps, the DTD method was shown to be instance-specific but not class-discriminative~\cite{gu2018understanding} -- the generated maps recognize foreground objects, but those maps are almost identical independently of the target class information. Following this, CLRP~\cite{gu2018understanding} and SGLRP~\cite{iwana2019explaining} have been developed in order to better specify the input features which trigger different output neurons. According to Gu \textit{et al.}~\shortcite{gu2018understanding}, the patterns of positively activated neurons are determined by (and hence also back-propagated to) the pixels on salient edges of the image, independently of the target class. The reverse propagation of different categories differs, however, in the exact relevance values that are assigned to the neurons. Based on that, the CLRP method utilizes the relative difference in relevance values of LRP maps of the same image for different categories as a tool for locating discriminative input features. SGLRP~\cite{iwana2019explaining} differs from the above CLRP method in the way it determines and distributes the initial relevance scores, where the post-softmax probabilities are used to weight non-target neurons in proportion to their contribution to the propagated heatmap rather than weighting them with a fixed value. Both CLRP and SGLRP methods result in class-discriminative features that mainly consist of contour lines that are relevant to the target class.  

We aim to harness the expressive nature of the above techniques for a better understanding of the influence of AEs on the network computational process. If these methods can reveal the input features that contribute to the decision made by the network, what will those features look like when the input is adversarial?

\subsection{Visualization Techniques and AEs}
\label{sec:vae}

Several studies have investigated the effect of AEs on visualization of features. Xu \textit{et al.}~\shortcite{xu2019interpreting} used different visualization techniques to provide pixel-wise analysis of the adversarial effects. Xiao \textit{et al.}~\shortcite{xiao2018spatially} used CAM to show how the attention of the network shifts to irrelevant regions of AE, and interpreted this as an indication of the attack's effectiveness. Dong \textit{et al.}~\shortcite{dong2017towards} used discrepancy maps to show how AEs differ from benign input, and proposed adversarial training scheme which incorporates consistency loss between true and adversarial maps in the objective function. The difference between adversarial and benign saliency maps was also explored by Gu and Tresp~\shortcite{gu2019saliency}. Finally, Chalasani \textit{et al.}~\shortcite{chalasani2018concise} show that adversarially trained networks tend to produce sparser and more concise explanations when using the Integrated Gradients method.

\section{LRP-based Methods and AEs}
\label{sec:lrp}

We present herein results of LRP-based methods on AEs, starting with some new insights achieved when using existing techniques, followed by the introduction of a novel LRP-based version.

\subsection{Class Discriminative Heatmaps of AEs}
\label{sec:cdheat}

\begin{figure}[h!]
\centering
\includegraphics[width=0.4\textwidth]{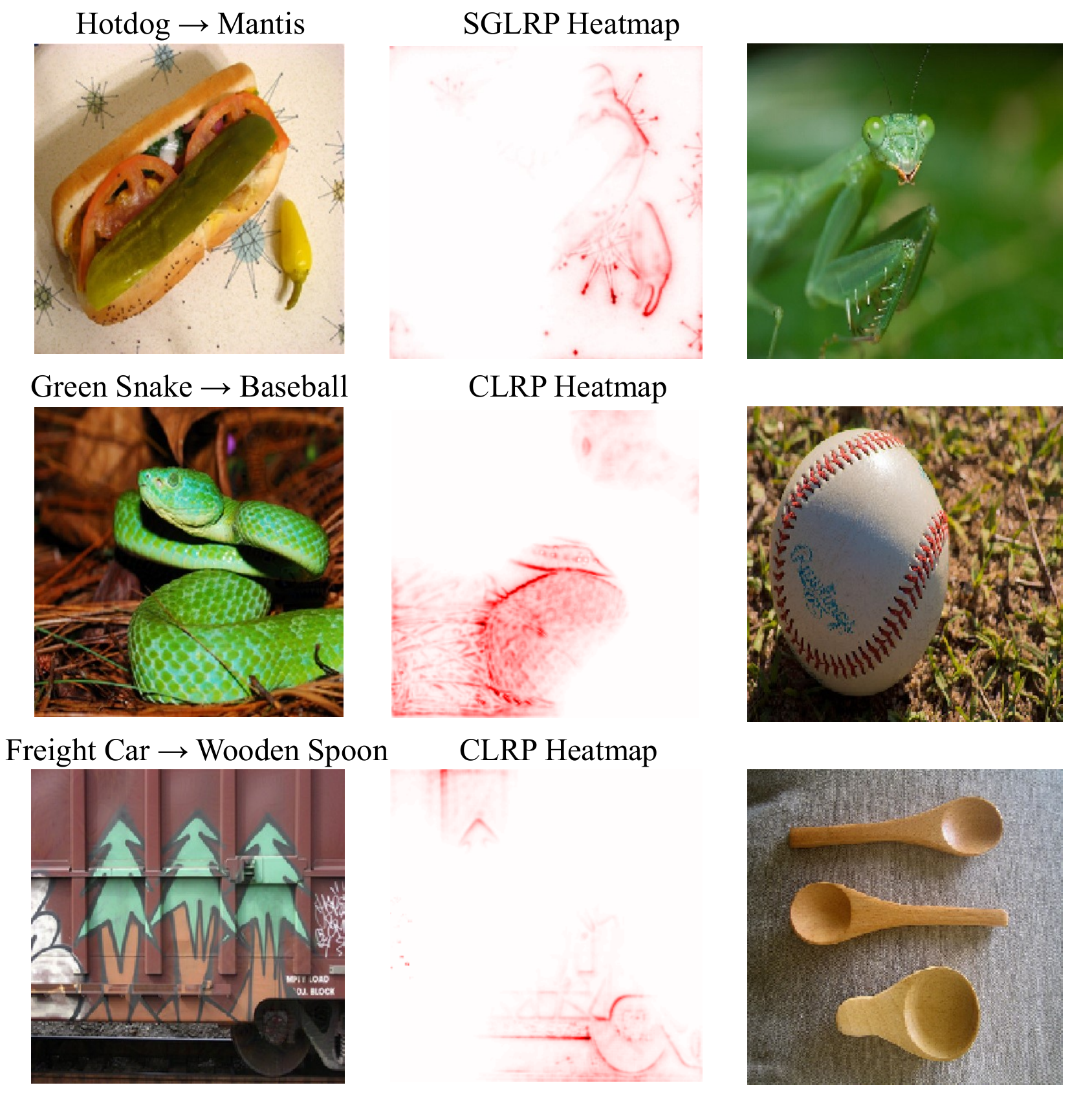}
\caption{Visualization of the adversarial effect on the network's decision neuron. The left column presents AEs created using Carlini-Wagner (CW) Attack (first row) and LBFGS Attack (second and third row). In the middle column are heatmaps produced for the predicted (adversarial) categories, and at the right column are examples from the ImageNet validation set of the adversarial objects.}
\label{fig:AEeffect}
\end{figure}

When applying class-discriminative methods to AEs, the effect of the attacks on the networks' decision making can be clearly visualized~\cite{dong2017towards,xiao2018spatially,gu2019saliency}. However, the previous works have either only stated that the focus of the heatmap changes as a result of an attack~\cite{xiao2018spatially,gu2019saliency}, or showed that for misclassification attacks, where the target class is very similar to the original classified class, the heatmap focuses more on features shared by both classes~\cite{dong2017towards}. We show here that the same phenomenon can surprisingly be observed also for strong attacks where the target class is very different from the original one. Figure~\ref{fig:AEeffect} presents three such examples, where the heatmaps of the adversarial images resemble objects from the adversarial target classes. These findings are not trivial, since one could speculate that for weak attacks, where misclassification is sufficient by itself, the attacks would enhance available features of an image that correspond to the non-primary classes. Strong target-class attacks, on the other hand, should require a different deceiving strategy, since the target class' features are not present in the original image. Explanations for such attacks, as was mentioned in Subsection~\ref{sec:ae}, usually involve more abstract and general aspects of NN computation such as ``unmapped islands'' in the input space. The current results suggest that for at least some of the strong attacks, the same strategy of enhancing non-target \emph{human-wise} features still applies.

 As was discussed by Iwana \textit{et al.}~\shortcite{iwana2019explaining}, the above methods do not always provide heatmaps as expected. Nevertheless, for the given examples, the adversarial effect can be interpreted as simply shifting of the network's attention from the principal features of the true object to the target (more peripheral) features. A more fundamental insight from this discussion is that the information about the different classes and their corresponding features exists in the inference process of both benign and adversarial inputs. The difference in the final decisions is a result of the way these features are weighted and prioritized.

\subsection{Absolute Value Heatmaps for Primary Class Specification}
\label{sec:avheat}

Motivated by the above observations, we hypothesize that the network preserves the knowledge of the ``true'' classification even though its actual output is incorrect. As stressed by Gu \textit{et al.}~\shortcite{gu2018understanding,gu2019saliency}, the class-specific information can be extracted from the saliency/relevance values assigned to the pixels rather than from the filtering effect itself. Previous works utilized this fact to generate class discriminative heatmaps by comparing relevance values between heatmaps of different categories. However, the fact that the discriminative pixels are relatively higher suggests that information about classes hides within the absolute relevance values as well.

\begin{figure}[h!]
\centering
\includegraphics[width=0.47\textwidth]{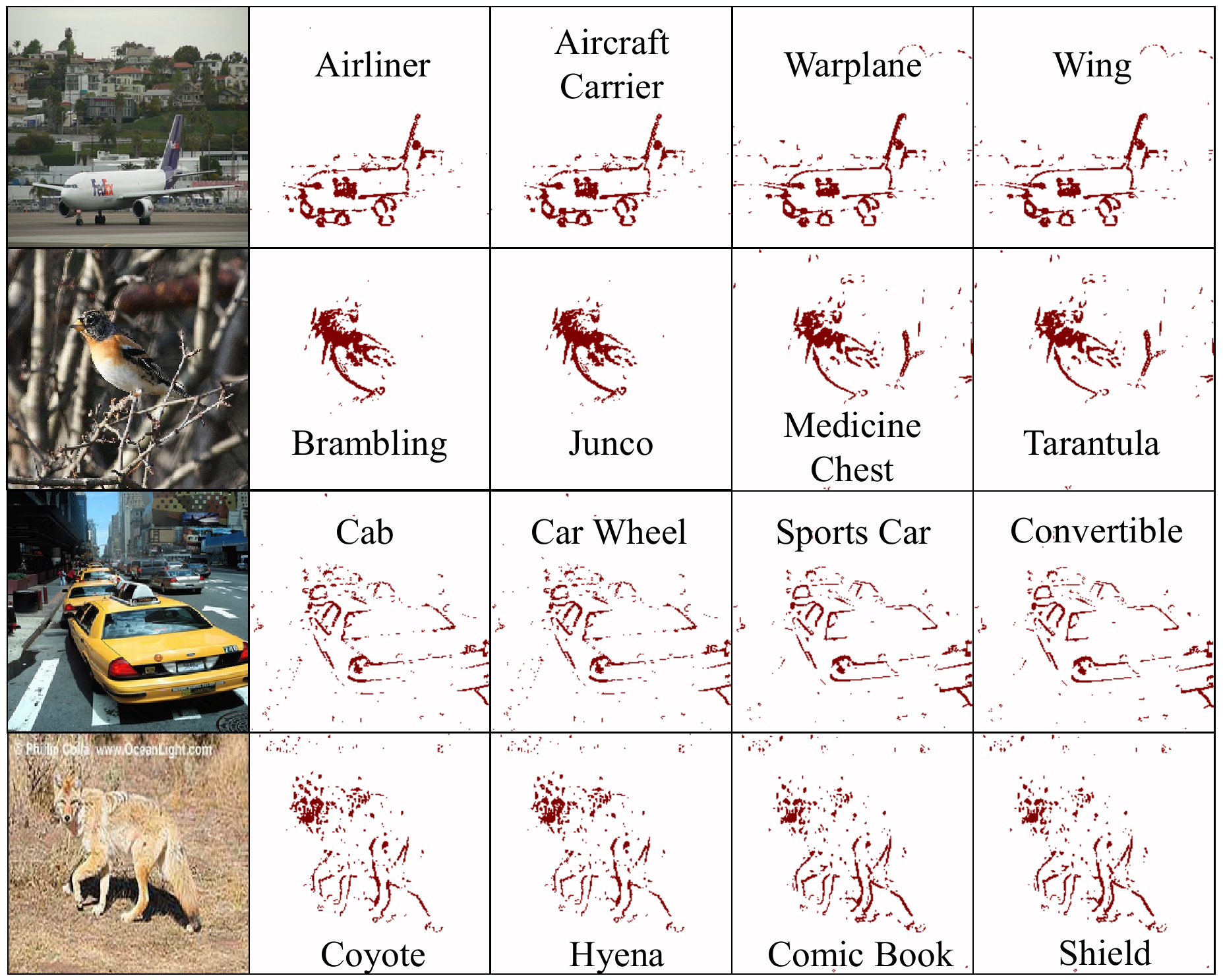}
\caption{Our new method produces similar heatmaps for both benign and adversarial categories. First column: AEs created using untargeted CW and targeted LBFGS attacks. Second and third columns: heatmaps for the original (benign) image's top-1 prediction and for another randomly selected prediction among the top-5 predictions, respectively. Forth and fifth columns: heatmaps for the AE's top-1 prediction and for another randomly selected prediction among the top-5 predictions, respectively.}
\label{fig:SimHmap}
\end{figure}

Specifically, we observe that the heatmap pixels with the highest absolute values, no matter which output neuron was that heatmap reverse-propagated for, belong to the primary object in the input image. This also applies \emph{even when the image is adversarial}; when reverse-propagating the relevance of the target class, the  pixels with the highest absolute relevance scores in the resulting heatmap still belong to the (true) primary object in that image. Figure~\ref{fig:SimHmap} presents how such primary class heatmaps are almost identical for different output neurons given some input image, as well as for the target neurons of different AEs which are based on that same image. The heatmaps were created by first reverse-propagating the one-hot encoding of the desired output neuron, using the standard DTD method~\cite{montavon2017explaining}, and then binarizing the pixels of the heatmap so that only the top 5\% of the pixels are preserved:
\begin{equation}
P_i =
\begin{cases}
    1 & P_i > \bar{p}+2 \cdot s_p\\
    0 & \text{otherwise}
\end{cases}
\label{eq3}
\end{equation}
where $\bar{p}$ and $s_p$ are the heatmap pixels' mean and standard deviation, respectively. 

\section{Heat and Blur}
\label{sec:heatblur}

Following the previous section, we now have a tool to highlight the pixels of the ``true'' objects in both benign and AEs. We suggest using this knowledge in the following ``heat and blur'' defense procedure, which is summarized in Algorithm~\ref{alg:heatblur}.

\begin{algorithm}[h!]
\caption{Heat and Blur}
\label{alg:heatblur}
\textbf{Input}: An image $i$\\
\textbf{Parameter}: Blurring factor $p_{blur}$\\
\textbf{Output}: A ``cleaned'' image $i_c$
\begin{algorithmic}[1] %[1] enables line numbers
\STATE Input the image to the model and get the output $y(i)$.
\STATE Obtain the heatmap $h_i =$ LRP$(y(i))$.
\STATE Use Equation~\ref{eq3} to get the binarized heatmap $b\_h_i$.
\STATE Blur the original image $i$ using a Gaussian filter with $\sigma=p_{blur}$ to obtain $i_c$.
\STATE Restore the original pixels that correspond to the entries of $b\_h_i$ and plant them in the blurred image $i_c$.
\STATE \textbf{return} $i_c$
\end{algorithmic}
\end{algorithm}

\begin{figure}[h!]
\centering
\includegraphics[width=0.47\textwidth]{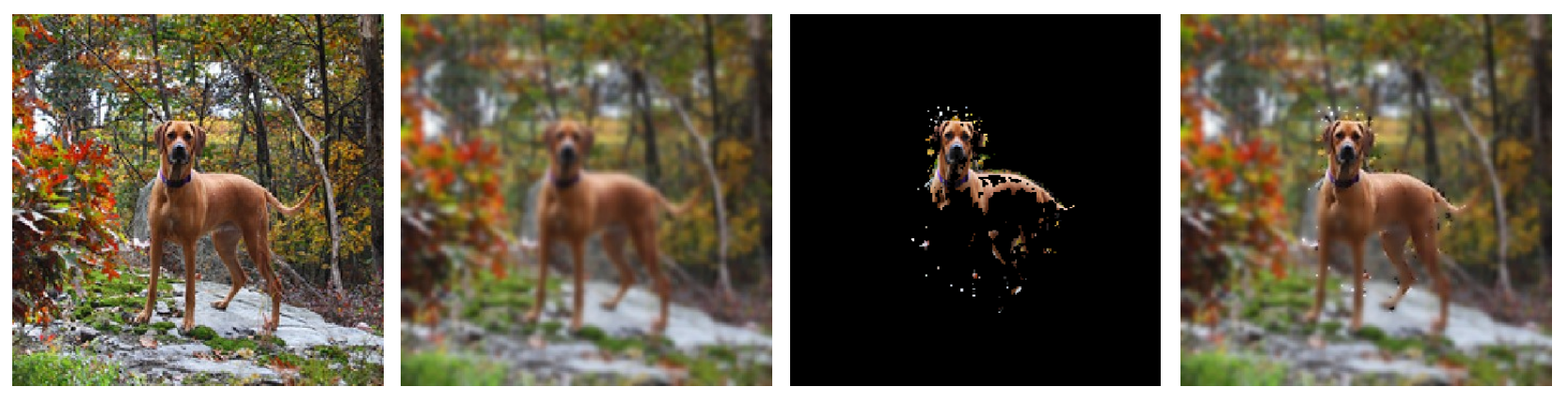}
\caption{Demonstration of the Heat and Blur method. From left to right: a) AE where `Rhodesian ridgeback' is misclassified as `Egyptian cat'; b) the blurred AE with $\sigma=1$; c) the recovered pixels according to the binarized heatmap; d) the `cleaned' image is now classified as `Rhodesian ridgeback'}
\label{fig:MethodDemo}
\end{figure}

In this method, we first input a suspicious image to a pre-trained model in order to obtain a heatmap of the primary object outlines, as was described in section 3.2. Then we significantly modify the image (in this case, using a Gaussian filter) to guarantee the cancellation of adversarial perturbations, which are by definition restricted in their amplitude in order to remain below human detection level. The resultant image is not adversarial, but may still be misclassified due to the massive distortion. Then, we restore the original values of the pixels that belong only to the heatmap from step 1. In this way, the contour lines of the "true" object are sharp and that object stands out, while the background remains blurry. It is important to preserve the background to some degree, since the texture and colors contribute to the decision making of the network, but the fine details are insignificant for the classification task. The modified image is then used as a new input for the final decision (see Figure~\ref{fig:MethodDemo} for a visual demonstration).

\begin{figure}[h!]
\centering
\includegraphics[width=0.48\textwidth]{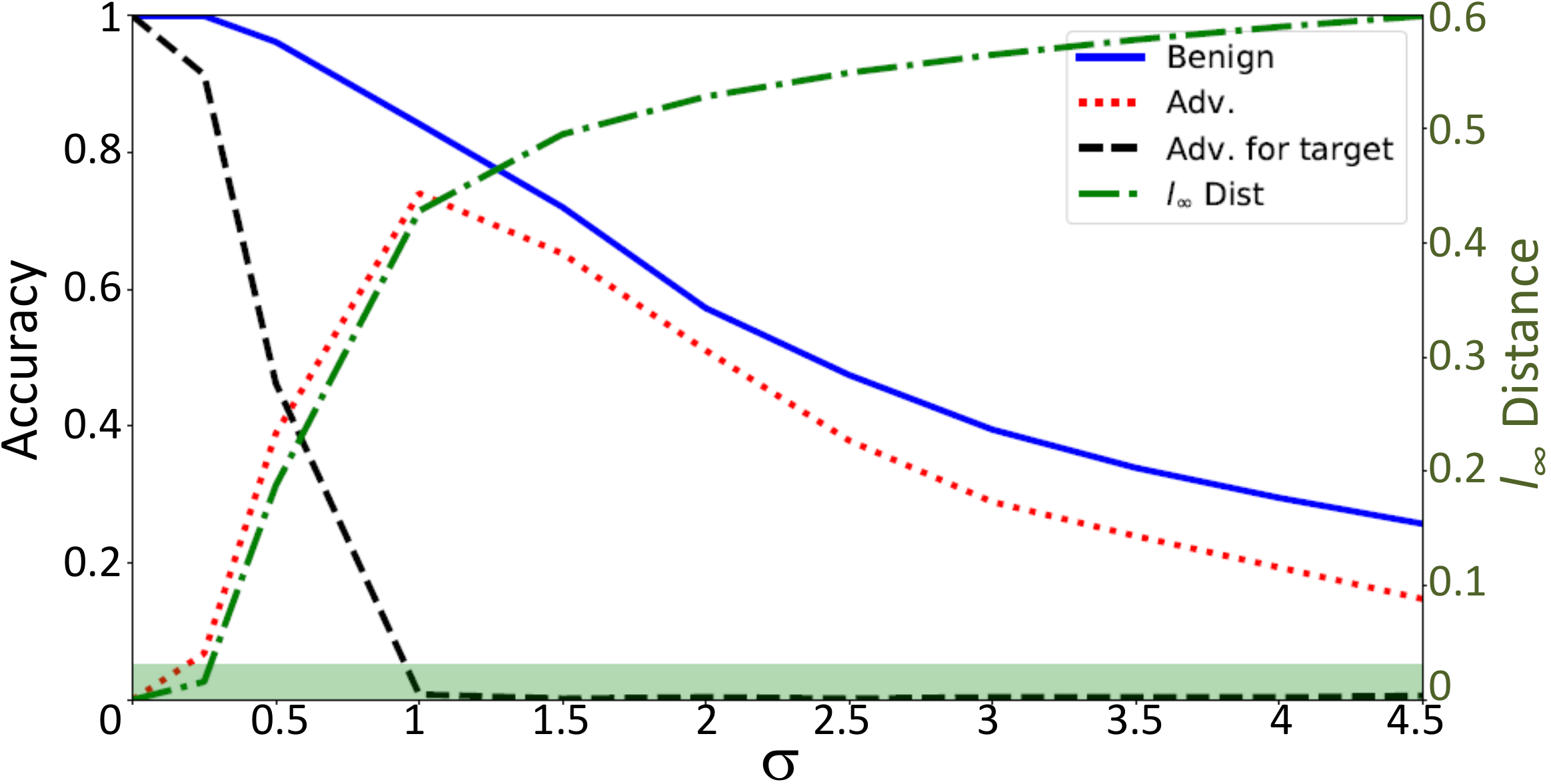}
\caption{Effect of the $\sigma$ blurring parameter on the accuracy of benign inputs (solid blue line), LBFGS AEs with respect to the true label (red dotted line) and adversarial terget category (black dashed line), and on the $l_{\infty}$ distance from the original input (green dash-dotted line) in the normalized [0, 1] space. The region of $l_{\infty}$ distance below the commonly used threshold of 0.031 is marked with green background}
\label{fig:BlurPar}
\end{figure}

We find that the above method is very efficient in ``clearing'' images from adversarial effects, with the cost of a moderate decrease in the network's performance on benign images. The balance between accuracy for benign (or cleaned) images and susceptibility to AEs is controlled by the $\sigma$ parameter, which determines the blurring degree (Figure~\ref{fig:BlurPar}). It requires two forward and one backward passes, which are computationally efficient when using designated platforms of deep learning. However, we recognize that using the traditional top-1 accuracy measure is not sufficiently informative as an evaluation metric.

\section{Evaluation Measurements}
\label{sec:evalmeasure}

Even though it is currently the most common measurement for the success of both attacks and defenses, top-1 accuracy answers the very limited question of ``did the NN classified the input as the exact correct label?''. This way, misclassification of `tiger shark' as `white shark' is equivalently wrong as misclassifying `tiger shark' as `broom', for example. Even the top-5 accuracy measure does not always capture the whole classification picture, since it only informs of whether the correct label appeared in those top-5 predictions, but does not indicate where it is located within those predictions, or how far it is from their semantic neighbourhood. There could be many applications where this distinction is important, for example if one wants to visually identify weapons in an image for security reasons, and an attacker masks weapons as animals. In this case, it does not matter if a defense strategy turns the adversarial `animal' into one kind of weapon or another, as long as the object is now identified as a threat. Alternatively, many attacks only switch between the top-$k$ (where $k$ is small) predictions of the model, which in many cases is not necessarily a wrong classification per say. Many images consist of several objects, for example an image labelled as `harvester' while there also appears `hay' in it. A top-1 classification of that image as `hay' provides true information regarding the content of that image, even though the correct label of `harvester' only appears as the second highest prediction due to the attack. On the other hand, other attacks may entirely change the semantic neighborhood of the top $k$ predictions, but this dramatic and more fundamental change is currently unaccounted for. To summarize, the current top-$k$ accuracy metric does not express the semantic and conceptual gap between the correct and adversarial predictions, while this may be valuable information in many use-cases.

We therefore suggest modifying a measure of the ranking quality that is widely used in Information Retrieval literature, called Normalized Discounted Cumulative Gain (NDCG)~\cite{croft2010search}. This measure was originally designed to apply graded relevance scores to results of search engines, based on the position of each element in the results list. We propose to use the pre- and post-softmax predictions of the model for a benign input as a source for relevance scores and as an ideal grading, by which attack and defense results should be normalized. Specifically, a set of post-softmax predictions $\sum_{j=1}^N p_j^{(i_b)}$ for a specific benign input $i_b$ will be scored as follows:
\begin{equation}
R_{p_j}^{(i_b)} =
\begin{cases}
 %   \frac{P_j^{(i_b)}}{K_1} & j\leq K_1\\
    \frac{l_j^{(i_b)}}{\sum_i l_i^{(i_b)}}             & j\leq K_1\\
    0                       & \text{otherwise}
\end{cases}
\label{eq4}
\end{equation}
where $l_j$ is the pre-softmax value of output neuron $j$, $K_1=\displaystyle\max_k \sum\nolimits_{j=1}^k p_j^{(i_b)}\leq C_b$ and $C_b =[0,1]$ is a parameter chosen as the prediction coverage. For example, for $C_b = 0.9$, only the $k$ highest predictions that sum up to 0.9 will be scored according to their original pre-softmax normalized value, and all the other $N-k$ predictions will be zeros. These scores will be used as relevance scores for the AE $i_a$ (or alternatively the cleaned image $i_c$) that was created on the basis of $i_b$:
\begin{equation}
R_{p_j}^{(i_a)} =
\begin{cases}
    R_{p_{j_{match}}}^{(i_b)} & j\leq K_2\\
    0                       & \text{otherwise}
\end{cases}
\label{eq5}
\end{equation}
where $p_{j_{match}}$ is the prediction within the $i_b$ output that corresponds to the category of $p_j$ (note that this prediction is not necessarily located in the $j$th place), and  $K_2=\displaystyle\max_k \sum\nolimits_{j=1}^k p_j^{(i_a)}\leq C_a$ with the parameter $C_a= [0,1]$.

The discounted cumulative gain of the predictions for an image $i$ is given by:
\begin{equation}
\text{DCG}_k^{(i)} = \sum_{j=1}^k \frac{2^{R_{p_j}^{(i)}}-1}{\log (1+j)}
\label{eq6}
\end{equation}
and the normalized DCG is defined as:
\begin{equation}
\text{NDCG}_K^{(i)} =  \frac{\text{DCG}_k^{(i)}}{\text{IDCG}}
\label{eq7}
\end{equation}
where we define IDCG as the ideal DCG given by the predictions for the benign image. Therefore the normalized DCG$_k$ for an AE $i_a$ is calculated as:
\begin{equation}
\text{NDCG}_{K_2}^{(i_a)} =  \frac{\text{DCG}_{K_2}^{(i_a)}}{\text{DCG}_{K_1}^{(i_b)}}
\label{eq8}
\end{equation}

\section{Experimental Evaluation}
\label{sec:exp}

\begin{figure*}[t!]
\centering
\includegraphics[width=1\textwidth]{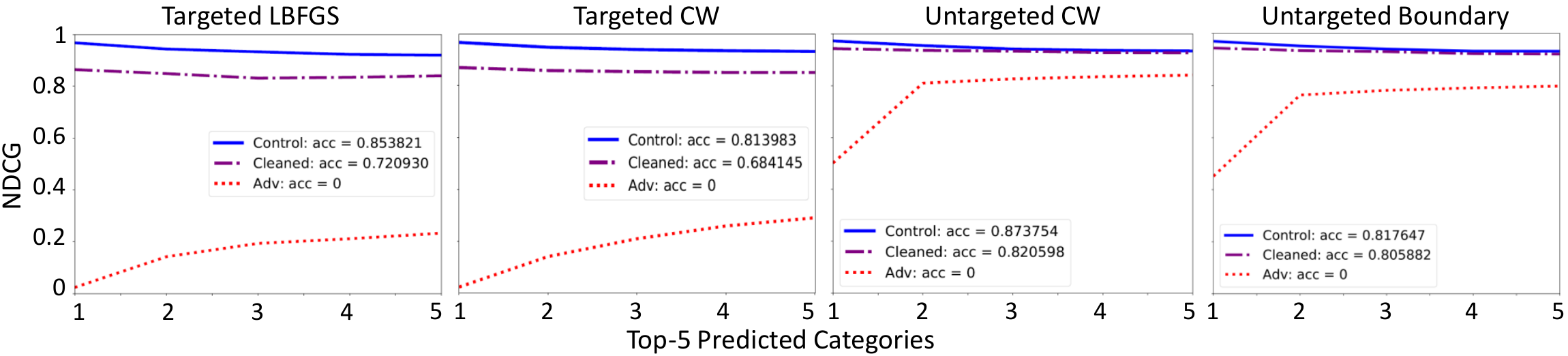}
\caption{Average NDCG scores for the 5 highest predictions of the model, compared between AEs (`Adv'), AEs after our method was applied (`Cleaned') and cleaned benign images (`Control'). Top-1 accuracy results for each group are presented in the legends.}
\label{fig:exp}
\end{figure*}

The experiments were conducted using VGG19 architecture with the ImageNet validation dataset. They include only benign images which were correctly classified by the model, and successfully applied AEs.  For each randomly selected image, an adversarial counterpart was created using the Foolbox\footnote{https://github.com/bethgelab/foolbox} toolbox~\cite{rauber2017foolbox}. The applied attacks consisted of $l_{2}$ bounded targeted and untargeted Carlini-Wagner (CW), targeted LBFGS and untargeted boundary attacks. For the targeted attacks, the target category was randomly selected, and the $l_{2}$ distance threshold was set to 0.02. The suggested defense procedure was based on the LRP method using the Innvestigate\footnote{https://github.com/albermax/innvestigate} toolbox~\cite{alber2019innvestigate}, and the threshold for the heatmap mask was empirically chosen to be the top 5\% of the LRP heatmap pixels. The blurring of the rest of the images was achieved using a Gaussian filter with $\sigma$ between [0.8, 1.2]. Each experimental group consisted of 500-800 couples of benign and adversarial images except for the boundary attack group with 175 couples, and the coverage parameter $C$ in the NDCG method was set to 0.99 for both benign and adversarial images. The heatmaps in Figure~\ref{fig:AEeffect} were created using the code provided by ~\cite{iwana2019explaining}\footnote{https://github.com/uchidalab/softmaxgradient-lrp}. 

\subsection{Evaluation of Heat and Blur as a Defense}
\label{sec:evaldefense}

We used the NDCG measure to evaluate the effectiveness of our suggested defense procedure against four different attack types. As can be observed in Figure~\ref{fig:exp}, the different attacks have a varying effect on the predictions of the model. For example, the top-1 accuracy of both targeted and untargeted AEs is 0, but these groups' top category NDCG scores are very different; while the targeted attacks cause the model to classify the AE as a random (and usually unrelated) class and hence the NDCG score is very close to 0, the untargeted attacks often directs the model to a category within the top-$k$ predictions of the original benign counterpart, as reflected in the NDCG score of $\sim$0.5. This higher score captures the fact that the misclassification is not always so dramatic and sometimes not even wrong, especially in cases where the input consists of several objects or can suffer several possible interpretations. The sharp increase of the untargeted AEs graphs is another indication of an attack strategy that often only switches the location of the top category, while the slow and incomplete rise of the targeted attacks' graph shows how these attacks completely change the relationships and semantic neighborhood of the top-$k$ categories. The effectiveness of our defense also varies, though it can be observed that it successfully recovers the correct context of the predictions, even when the top-1 accuracy is limited. Also note that for the control benign group, the NDCG scores are higher and relatively consistent across the different experimental groups, whereas the top-1 accuracy measure is more sensitive to noise. 

\subsection{Adaptive Attacks}
\label{sec:evalattack}

Since some of the pixels remain untouched, one can argue that an adaptive attack is still possible, where perturbations are only applied to the heatmap pixels of the ``true'' object. In this way, the defense will fail to clear the adversarial effect and the network will still misclassify the image. In order to explore this possibility, we created a set of new attacks that indeed consist of perturbations to the top-5\%  pixels alone (see Algorithm~\ref{alg:adaptive} for the detailed procedure). We used several attack types such as CW or PGD, and found that constraining the possible perturbations to such a small portion of the image pixels has a dramatic impact on the attack process. If the perturbation size is not strictly defined, the resultant perturbations are above the human detection threshold. When trying to apply attacks with $l_{\infty}$- or $l_2$-norm bounded perturbations, the algorithm usually fails to find an attack (we have succeeded in less than 1\% of the attempts). In addition, all the AEs that we have managed to produce have been (wrongly) classified by the network with low confidence and into a semantically related class. We believe that further refinements of our suggested defense, combined with more thorough analysis of the vulnerable pixels that belong to the heatmap, will help circumvent even the few currently successful attacks.

%\begin{figure*}[t!]
%\centering
%\includegraphics[width=0.67\textwidth]{fig5.jpg}
%\caption{Adaptive attacks. The top leftmost AE is the only attack with $l_\infty$ norm below human detection (0.0125) that we have managed to generate (and the adversarial class is very similar to the original one), using a modified version of Basic Iterative Attack. All the other AEs contain observable pixel changes ($l_\infty$ norm above 0.03). The two bottom images were generated based on PGD attack with $l_\infty$ norm and LBFGS attack with $l_2$ norm, respectively.}
%\label{fig:adaptive}
%\end{figure*}

\begin{algorithm}[h!]
\caption{Adaptive Attack}
\label{alg:adaptive}
\textbf{Input}: An image $i$, an attack method Att(.)\\
\textbf{Parameter}: Number of iterations $iter$\\
\textbf{Output}: An adaptive AE $i_{adapt}$ or $\emptyset$
\begin{algorithmic}[1] %[1] enables line numbers
\STATE $i_{temp} = i$
\FOR{$iter$ iterations}
    \STATE $i_{adv}$ = Att($i_{temp}$)
    \STATE hmap\_mask = binarized\_heatmap($i_{temp}$)
    \STATE $i_{temp}$[hmap\_mask] = $i_{adv}$[hmap\_mask]
    \STATE $i_{adapt}$ = clip($i_{temp}$) to be in the allowed pixel range
    \IF {$i_{adapt}$ is adversarial}
        \STATE \textbf{return} $i_{adapt}$
    \ELSE
        \STATE $i_{temp}$ = $i_{adapt}$
    \ENDIF
\ENDFOR
\STATE \textbf{return} $\emptyset$
\end{algorithmic}
\end{algorithm}

\section{Discussion and Conclusions}
\label{sec:conc}

The above results exemplify how interpretation methods such as LRP can be used for defense against AEs. We chose to focus on LRP since this method is computationally efficient and conceptually straightforward, and does not require many assumptions and hyperparameters tuning. Using a Gaussian filter for the defensive distortion is not essential and other types of randomized noise or input transformations may give comparable or even better results, as the connection between adversarial and corruption robustness was thoroughly explored by Ford \textit{et al.}~\shortcite{ford2019adversarial}. The choice of the 5\% highest pixels for localization of the primary object, though empirically effective, has some limitations. The relevance heatmaps are sometimes rather sparse, especially in cases where the model predictions are vague, and as a result the top 5\% pixels may be too diluted. A more sophisticated determination of the percentage of pixels to restore may improve the classification accuracy of those cases. 
Overall, the simplicity of the suggested method could enhance its robustness, as will be discussed below, and it could also facilitate integration with other defenses. For example, scanning of the restored pixels for statistically atypical values, or defending against possible attacks on the interpretation method itself.

Recent studies have demonstrated that interpretability methods may also be vulnerable to adversarial attacks. For example, it was shown that saliency methods are sometimes sensitive to input transformations such as adding a constant to the input~\cite{kindermans2019unreliability}, however the results are less relevant to our defense since the highest-valued pixels of the attacked heatmaps seem to remain intact. In the work of Zhang \textit{et al.}~\shortcite{zhang2018interpretable}, it was shown that pixel manipulation can indeed produce wrong interpretation heatmaps, but their method was not tested against LRP-based methods. The authors have, however, showed that the transferability of AEs across different interpreters is low, a fact that makes black-box attacks more difficult than standard classification attacks. They also presented defense methods against such attacks. Ghorbani \textit{et al.}~\shortcite{ghorbani2019interpretation} show that several interpretation methods are fragile to small perturbations, which change the produced heatmaps while maintaining the same predictions. They explain this vulnerability as a result of the high dimensionality and non-linearities in deep networks which affect gradient-based interpretations; intuitively, two sets of classes can be separated similarly using different boundaries, so that the gradients may change and lead to misleading interpretations. Since their attack method was only tested against gradient-based interpretations where the prediction of the model was intact, it is left for future research to determine whether methods like our LRP version  that are not based directly on gradients are vulnerable as well. If so, then new intuition and explanation are required for understanding the effect of such attacks, as well as new defense strategies.

\section*{Acknowledgements}
\label{sec:ack}

This research was supported by the Ministry of Science \& Technology, Israel, and by the Ariel Cyber Innovation Center in conjunction with the Israel National Cyber directorate in the Prime Minister's Office.

%We also ran experiments on FGSM and PGD attacks, which show similar trends to those depicted in Figure~\ref{fig:exp}, though with lower top-1 accuracy scores (0.588040 and 0.???, respectively).

\small
\bibliographystyle{named}
\bibliography{NN}

\end{document}